\documentclass[conference]{IEEEtran}
\IEEEoverridecommandlockouts
\usepackage{cite}
\usepackage{amsmath,amssymb,amsfonts}
\usepackage{algorithmic}
\usepackage{graphicx}
\usepackage{subfigure}
\usepackage{multirow}
\usepackage{textcomp}
\usepackage{xcolor}
\def\BibTeX{{\rm B\kern-.05em{\sc i\kern-.025em b}\kern-.08em
    T\kern-.1667em\lower.7ex\hbox{E}\kern-.125emX}}
\usepackage[a4paper,top=54pt,bottom=113pt,left=37pt,right=37pt]{geometry}
\begin{document}

\title{A Local Descriptor with Physiological Characteristic  for Finger Vein Recognition\\
}

\author{\IEEEauthorblockN{Liping Zhang, Weijun Li, Xin Ning}
\IEEEauthorblockA{\textit{Institute of Semiconductors, Chinese Academy of Sciences} \\
\textit{Beijing Key Laboratory of Semiconductor Neural Network Intelligent Sensing and Computing Technology} \\
\textit{School of Microelectronics, University of Chinese Academy of Sciences} \\	
\textit{Cognitive Computing Technology Joint Laboratory, Wave Group}\\
Beijing, China \\
\{zliping, wjli, ningxin\}@semi.ac.cn}

}

\maketitle

\begin{abstract}
Local feature descriptors exhibit great superiority in finger vein recognition due to their stability and robustness against local changes in images. However, most of these are methods use general-purpose descriptors that do not consider finger vein-specific features. In this work, we propose a finger vein-specific local feature descriptors based physiological characteristic of finger vein patterns, i.e., histogram of oriented physiological Gabor responses (HOPGR), for finger vein recognition. First, prior of directional characteristic of finger vein patterns is obtained in an unsupervised manner. Then the physiological Gabor filter banks are set up based on the prior information to extracte the physiological responses and orientation. Finally, to make feature has robustness against local changes in images, histogram is generated as output by dividing the image into non-overlapping cells and overlapping blocks. Extensive experimental results on several databases clearly demonstrate that the proposed method outperforms most current state-of-the-art finger vein recognition methods.
\end{abstract}


\section{Introduction}
Finger vein recognition is one of the representative techniques for human identification. It has many advantages, such as non-contact collection, vivo identification, internal features, small size, and simple structure of imaging devices. Due to these positive properties, finger vein recognition has attracted considerable attentions in the field of biometric recognition \cite{jain2004introduction, kumar2011human}.

Many efforts have contributed to the development of finger vein recognition, and the technology has made remarkable progress in the last two decades \cite{miura2004feature, yang2018finger, liu2017finger, xi2017learning, meng2018finger}. Repeated line tracking (RLT) \cite{miura2004feature} extracts vein patterns from multiple random starting positions. Principal component analysis (PCA) \cite{wu2011finger} and linear discriminant analysis (LDA) \cite{wu2011} are two representative global feature-based methods that learn feature mapping to retain the statistical information of finger vein images. Local binary pattern (LBP) \cite{lee2009finger} is a typical local feature descriptor that compares the intensity of a central pixel to its neighbor pixels. Guided filter-based single scale retinex (GFSSR) \cite{xie2015intensity} has been proposed as an intensity variation normalization method for finger vein recognition. Discriminative binary codes (DBC) \cite{xi2017learning} learning is a finger vein recognition method, in which subject relationship is considered in order to improve performance. Kang et al. \cite{kang2018noise} presented a method that exploits the intensity distribution as a novel soft biometric trait and combines the primary and soft biometric traits on the score level to achieve better performance. However, in these methods, if the captured images have low contrast, it is difficult to achieve accurate vein pattern segmentation, and the recognition performance of vein pattern-based methods degrades dramatically. Global feature mapping usually lacks robustness in local intra-class variations that arise due to variations in posture, rotation, uneven illumination, scale, etc.

2D Gabor filter is widely used in various visual recognition applications and has exhibited desirable performance in capturing texture and orientation information \cite{rahman,luan2018gabor,jia2013histogram}, as well as played an important role in finger vein recognition technology. Based Gabor filter, Gabor wavelet responses (GWRs) \cite{kumar2011human}, Gabor+Tri-branch structure \cite{yang2017tri}, point grouping assisted Gabor (PG-Gabor) \cite{yang2019point}, and adaptive learning Gabor filters (ALGF) \cite{zhang2019adaptive} are all proposed for finger vein recognition. However, all previous finger vein recognition methods based on Gabor filters use a fixed set of filters with orientations obtained via constant sampling in the range ${\left[{0, \pi}\right]}$ and fixed values of scale and frequency\cite{genovese2019palmnet}.

Among all finger vein recognition methods, local feature-based methods exhibit great superiority due to their stability and robustness in relation to local changes in feature description. As a local feature-based method, histogram of oriented gradients (HOG) captures the edge or gradient information from a local region, and it is not very sensitive to changes in brightness, It can withstand small transformations as the histogram values  are only minimally affected by small translations and rotations \cite{dalal2005histograms}. However, gradient information used in HOG is not a good representation of finger vein patterns, because the patterns have different widths and crisscrossing vein patterns. As variants of HOG, histogram of competitive Gabor responses (HCGR) \cite{lu2014finger} and histogram of oriented lines (HOL) \cite{Zhangliping2018} use Gabor wavelets to detect the line responses and orientations of pixels. However, these methods use general-purpose filters that do not consider finger vein-specific features, which limits the adaptability and efficiency of the filters for finger vein images.

In this work, through the analysis of the physiological characteristic of finger veins patterns, we attempt to probe an effective technique that can exploit the rich oriented features in finger vein images. A novel local feature descriptor called histogram of oriented physiological Gabor responses (HOPGR) is proposed in the present study for finger vein recognition. The proposal takes advantage of the 2DGabor filter’s ability to capture texture from setting orientations and extract histogram feature of the finger vein images. Specifically, we combine Gabor filters with the physiological characteristic of finger vein patterns and propose a new method of filter direction setting to replace the constant sampling way, which is more suitable for finger vein patterns. We obtain physiological information from a finger vein dataset containing images of fingers collected from multiple individuals. This is instrumental in creating the accurate representation of the orientation and local texture in a finger vein image. Several experimental results using two databases demonstrate the effectiveness of the proposed method for finger vein recognition.

\section{THE PROPOSED METHOD}
\subsection{Overview}
As illustrated in Fig. 1, The main framework of the HOPGR can be divided into two phases, namely prior information acquisition and HOPGR feature extraction. In the first phase, the directional characteristic of finger vein patterns is analysed, and the common rules from a training dataset is obtained as prior information in an unsupervised manner. In the second phase, the physiological Gabor filter banks are set up based on the prior information, the responses and orientation are then extracted though the physiological Gabor filter banks. Finally, to make feature has the robustness against local changes in images, histogram is generated as output by dividing the image into non-overlapping cells and overlapping blocks.

\begin{figure}[t]
	\centering
	\includegraphics[width=3.5in]{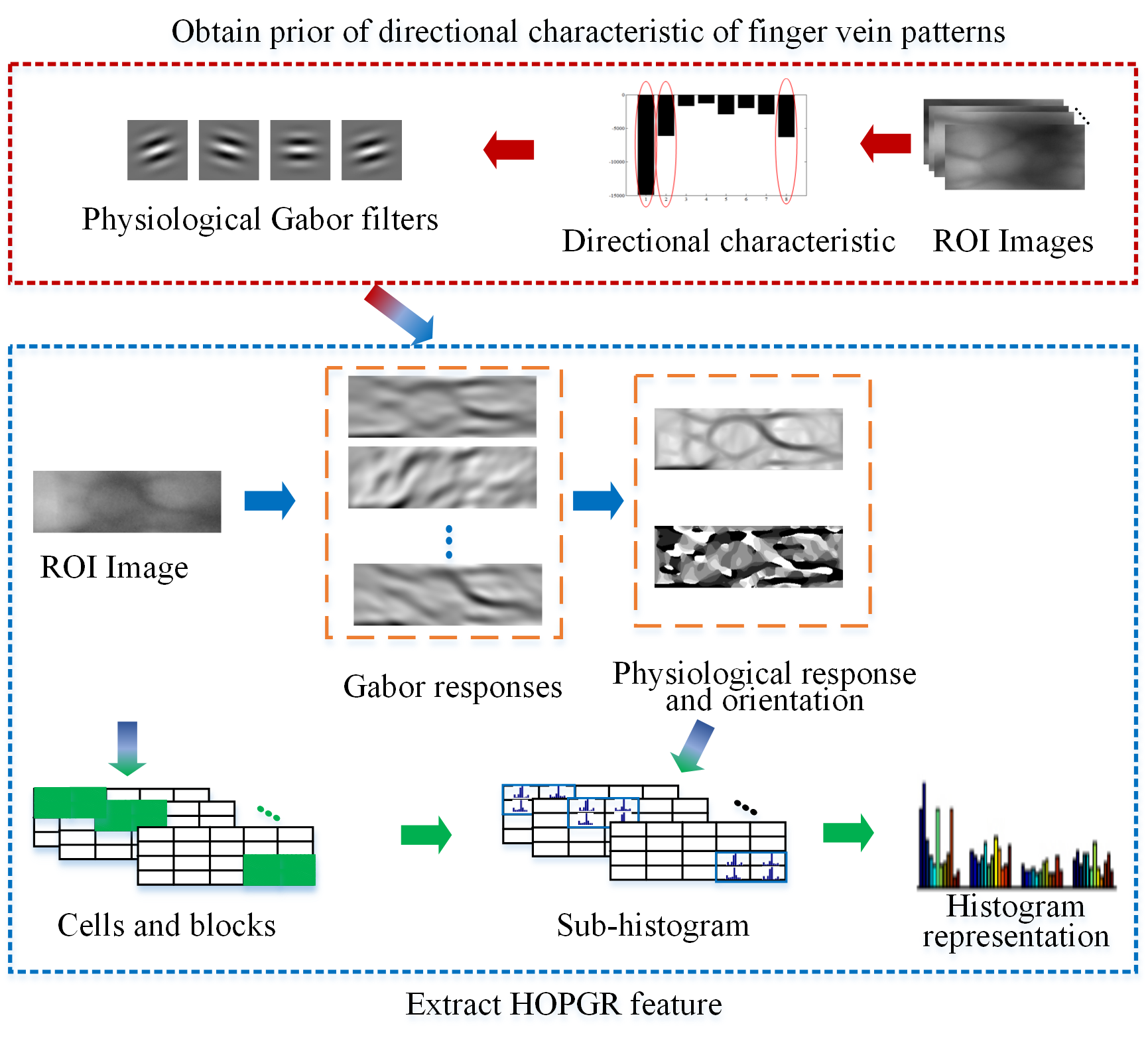}
	\caption{Illustration of the HOPGR descriptor. The main framework consists of two phases: prior information acquisition phase and HOPGR feature extraction phase.}
\end{figure}

\subsection{Physiological Gabor Responses of Finger Vein Image}
\begin{figure}[t]
	\centering
	\includegraphics[width=3in]{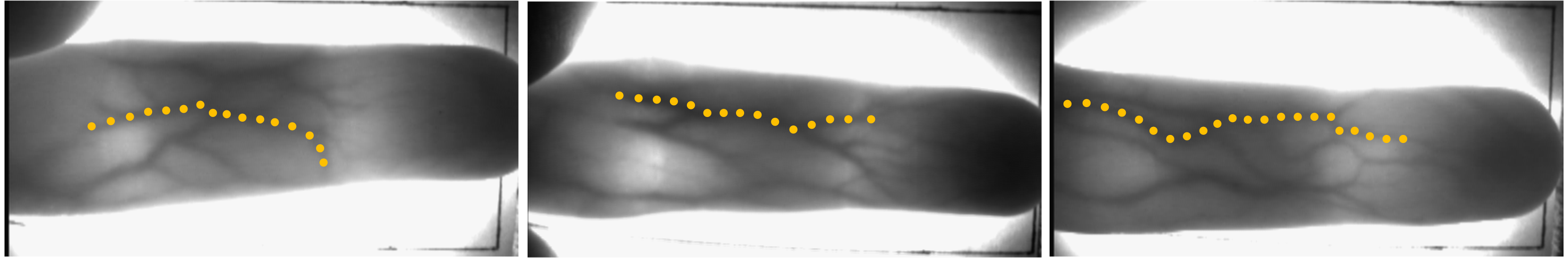}
	\caption{Some examples showing physiological characteristic of finger veins that grow along the direction of the finger.}
\end{figure}

Gabor function can be a powerful line-shape filter for extracting texture features of a finger vein in the set directions, and it is defined as follows:
\begin{equation}
G(x,y,\theta ,{f_0},\sigma )\\
= \frac{1}{{2\pi {\sigma ^2}}}\exp \left[ {\frac{{-1}}{2}\left( {\frac{{x_\theta ^2 + y_\theta ^2}}{{{\sigma ^2}}}} \right)} \right]\cos (2\pi {f_0}{x_\theta })
\end{equation}
where ${{x_\theta }{\rm{ = }}x\cos \theta  + y\sin \theta ,\;{y_\theta }{\rm{ = }} - x\sin \theta  + y\cos \theta}$, ${\theta}$ is the orientation of the Gabor function, ${{f_0}}$  denotes the central frequency of the sinusoidal wave, and ${\sigma}$  represents the standard deviation of the Gaussian envelope. The direction ${{\theta_k}}$ is calculated \cite{lu2014finger} as follows:
\begin{equation}
{\theta _k} = \frac{{\pi (k - 1)}}{O},k = 1,2, \cdots ,O
\end{equation}
where ${O}$ is the quantity of orientations. The more orientations are set, the stronger is the ability of the filter bank to obtain the texture information. However, more orientations set leads to a corresponding increase in computational costs.

Fig. 2 shows examples of finger vein images with labeled vein patterns. As can be seen from the figure, finger vein patterns have clear directional distribution. Specifically, along the length of the finger, and if the finger is pointed horizontally, the veins distribute more in the near-horizontal orientation and less in other orientation.

\begin{figure*}[t]
	\centering
	\includegraphics[width=5in]{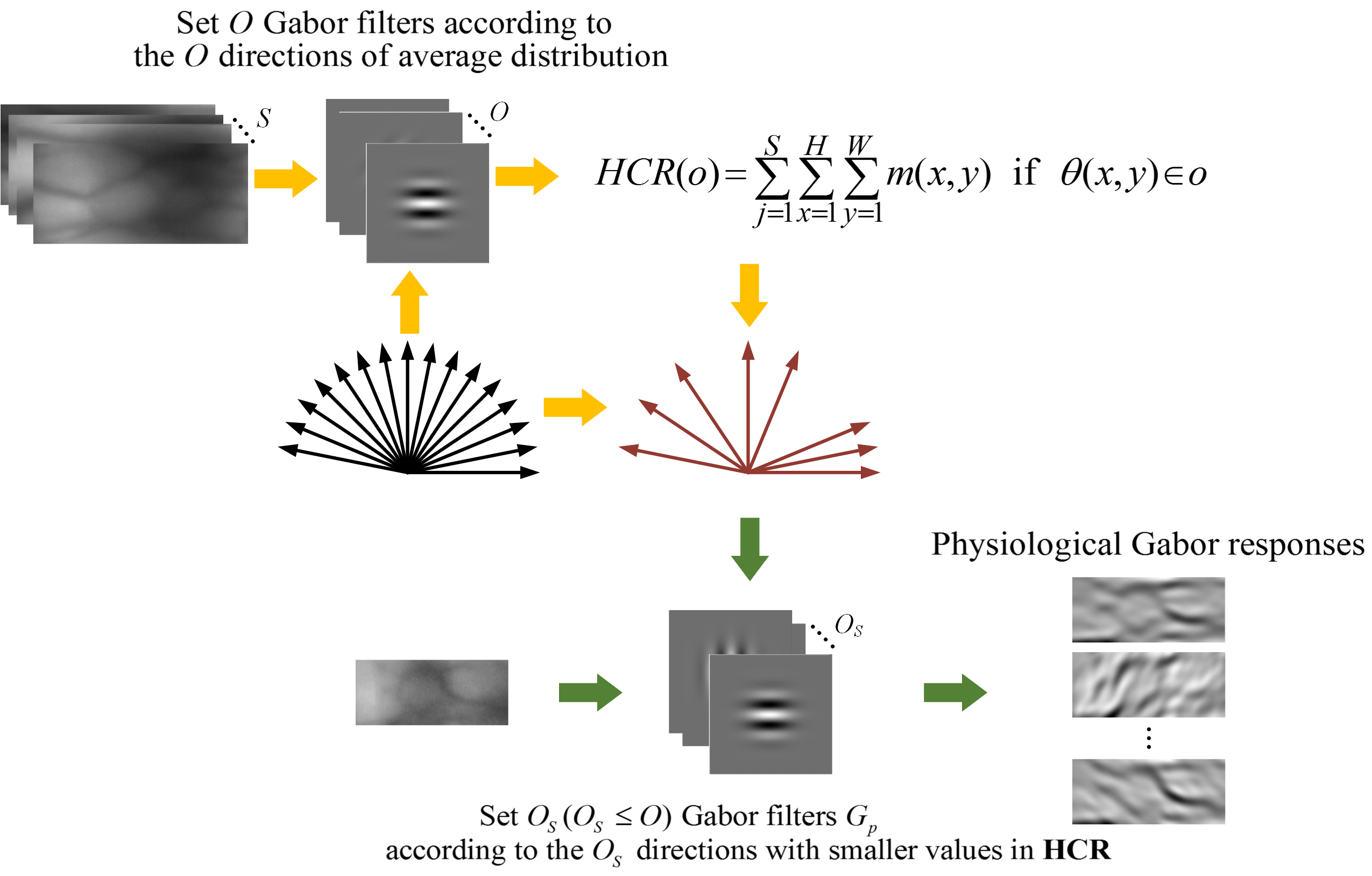}
	\caption{The flow-chart for extracting physiological Gabor responses of finger vein image. During the phase of obtaining prior physiological information, the common rules from a dataset with ${S}$ finger vein images are summarized. The values in ${{\bf{HCR}}}$ with directions of average distribution are computed and then ${{O_S}({O_S} \le O)}$ directions with smaller values in ${{\bf{HCR}}}$ are selected. In application phase, the Gabor filters are set with the selected directions to obtain the physiological responses.}
	
\end{figure*}
Taking into account physiological characteristic, a new way of direction setting for Gabor filters  is proposed in order to extract vein patterns more effectively. Fig. 3 shows the details.

To use this specific characteristic, the explicit and common rules from multiple fingers of multiple people are summarized first. Let ${{\bf{Y}} = [{{\bf{y}}_1};{{\bf{y}}_2}; \cdots ;{{\bf{y}}_S}]}$  be the ${S}$ finger vein images that provide physiological information. First, for each {\bf{y}} enough equational orientations are set by Eq. 2, to obtain the Gabor responses of each pixel ${y(x_1,x_2)}$ in multiple directions by convolution.
\begin{equation}
{F_k}(x,y){\rm{ = }}y(x_1,x_2) * G
\end{equation}

Based on ${{F_k}(x,y)}$ , the response  ${m{(x,y)}}$ and orientation ${\theta {(x,y)}}$  are computed in the following way:
\begin{equation}
m{(x,y)} = \min ({F_k}(x,y))
\end{equation}
\begin{equation}
\theta {(x,y)} = \arg \min ({F_k}(x,y))
\end{equation}

Then a histogram of cumulative responses (${{\bf{HCR}}}$) of all pixels in all ${S}$ finger vein images is calculated as follow:

\begin{equation}
HCR(o) = \sum\limits_{j = 1}^S {\sum\limits_{x = 1}^H {\sum\limits_{y = 1}^W {m(x,y)} } } {\rm{\ if \ }}\theta (x,y) \in o
\end{equation}
where ${W}$ and ${H}$ are respectively the width and height of a finger vein image.

The vein patterns depicted in a finger vein image are composed of lines darker than the brightness value of the non-vein area \cite{miura2004feature}. Therefore, there is a greater distribution of veins in the directions with smaller values in ${{\bf{HCR}}}$ . In view of this, ${{O_s}({O_s} \le O)}$ directions with smaller values in ${{\bf{HCR}}}$  are ultimately selected as the directions of the Gabor filter banks. As such, let ${G_p}$ be the filters that are set in selected directions. 

During the application phase, utilizing ${G_p}$, which conforms to the physiological characteristic of finger vein patterns, to obtain the physiological Gabor responses of finger vein image ${I}$ as follows:
\begin{equation}
{F_s}(x,y){\rm{ = }}I(x,y) * {G_p}, s\in {O_S}
\end{equation}

The physiological response ${m{(x,y)}}$ and orientation ${\theta {(x,y)}}$ are computed as:
\begin{equation}
{m_p}{(x,y)} = \min ({F_s}(x,y))
\end{equation}
\begin{equation}
{\theta_p} {(x,y)} = \arg \min ({F_s}(x,y))
\end{equation}

\subsection{Histogram Feature Generation}
To make feature has the robustness against local changes in images, histogram is constructed as output by dividing the image into non-overlapping cells and overlapping blocks.

\textbf{Step 1.} For each image ${I}$, Non-overlapping cells and overlapping blocks are constructed as illustrated in Fig. 1.

\textbf{Step 2.} For each cell, the sub-histogram within the cell ${{\bf{HC}}}$ is calculated based on the orientation, and each sub-histogram has ${O}$ bins:
\begin{equation}
HC{(o)_i} = \sum {{m_p}(x,y)} \\
{\rm{\ if \ }}I(x,y) \in cel{l_i}{\rm{ }}\ and\ {\theta_p} (x,y) \in {\rm{bin}}(o)
\end{equation}

\textbf{Step 3.} For each block, the histogram of the block ${{\bf{HB}}}$ can be obtained by combining all the ${{\bf{HC}}}$s within the block:

\begin{equation}
\bf{H{B_j}} = \left\{ {H{C_1},H{C_2}, \cdots ,H{C_{{b_1} \times {b_2}}}} \right\}
\end{equation}
Generally, a block consists of ${2 \times 2}$ cells.

\textbf{Step 4.} Each ${{\bf{HB}}}$  is normalized as follows via \emph{L2-norm} using a small constant ${\varepsilon}$  to avoid division by zero:

\begin{equation}
{\bf{NH{B_j}}}   {\rm{ = }}\frac{{\bf{H{B_j}}}}{{\sqrt {\left\| {{\bf{H{B_j}}}} \right\|_2^2{\rm{ + }}{\varepsilon ^2}} }}
\end{equation}

${{\bf{NHB}}}$s are concatenated together to generate the HOPGR feature in a vector form.

\section{Experiments}

\subsection{Databases}	
\begin{figure}[t]
	\centering
	\includegraphics[width=3in]{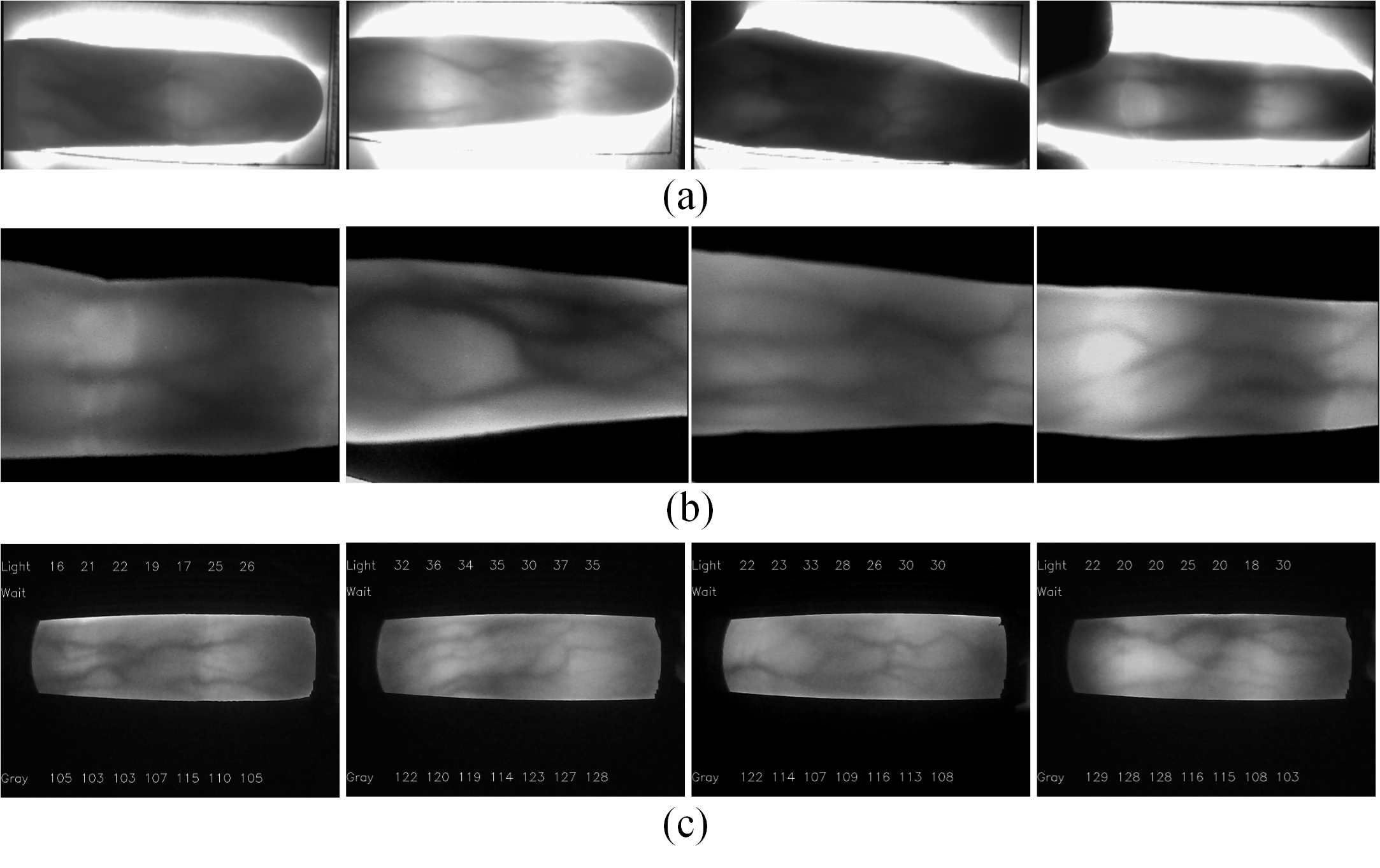}
	\caption{Some samples of the finger vein databases: (a) HKPU database, (b) MMVBNU\_6000 database, (c) SEMI-FV database.}
\end{figure}
In our paper, the open finger vein database HKPU \cite{kumar2011human} with the largest number of subjects was employed to provide prior information on physiological characteristic . All of the images captured in the first and second sessions of HKPU were used, totaling 3,132 images. The open database MMVBNU\_6000 and a self-built finger vein database SEMI-FV were used for performance evaluation.  MMVBNU\_6000 database \cite{lu2013available} was constructed by Chonbuk National University. It includes 6000 images of 100 volunteers; 10 images of 6 fingers of each volunteer were collected. The SEMI-FV database \cite{Zhangliping2018} is a self-built finger vein database that consists of 2610 finger vein images captured from 29 volunteers, with images of six fingers per subject. Some image samples are depicted in Fig. 4.

Region of interest (ROI) images from the MMVBNU\_6000 database were adopted directly \cite{lu2013robust}. ROI images from the HKPU and SEMI-FV databases, sized ${370 \times 130}$, were extracted using a preprocessing method similar to that described in the prior work \cite{lu2013robust}. In all the experiments, the performance of the methods is evaluated by the equal error rate (EER), which is the error rate when the false rejection rate (FRR) is equal to the false acceptance rate (FAR).

\subsection{Analysis of the Prior Physiological Characteristic}

To ensure reliability of the statistical information and versatility of the directional characteristic, the entire HKPU database was employed to provide physiological direction information. Therein, ${O = 16}$  was first set, and then directions were calculated as ${\theta  = 0{\rm{,}}\frac{\pi }{{16}}{\rm{,}} \cdots {\rm{,}}\frac{{15\pi }}{{16}}}$ according to Eq. 2. The ${{\bf{HCR}}}$ were computed according to Eq. 6 and are presented in Fig. 5. As illustrated, the finger veins were found to have unique directional characteristic. The values in all 16 directions in ${{\bf{HCR}}}$ exhibited great dissimilarity. The values in ${{\bf{HCR}}}$  were then sorted in an ascending order, and the results are presented in Table I.

It is clear from the results that the value in the horizontal direction (${\theta  = 0}$) was more than 89 times of that in (${\theta  = \frac{{5\pi }}{8}}$). So the first 8 orientations(${\theta  = 0, \frac{{\pi }}{16}, \frac{{\pi }}{8}, \frac{{3\pi }}{16}, \frac{{\pi }}{2}, \frac{{13\pi }}{16}, \frac{{7\pi }}{8}, \frac{{15\pi }}{16}}$) with smaller values were then selected as directions of Gabor filters to obtain the physiological responses. The selected directions are shown in Fig. 6.

\begin{figure}[tb]
	\centering
	\includegraphics[width=2.5in]{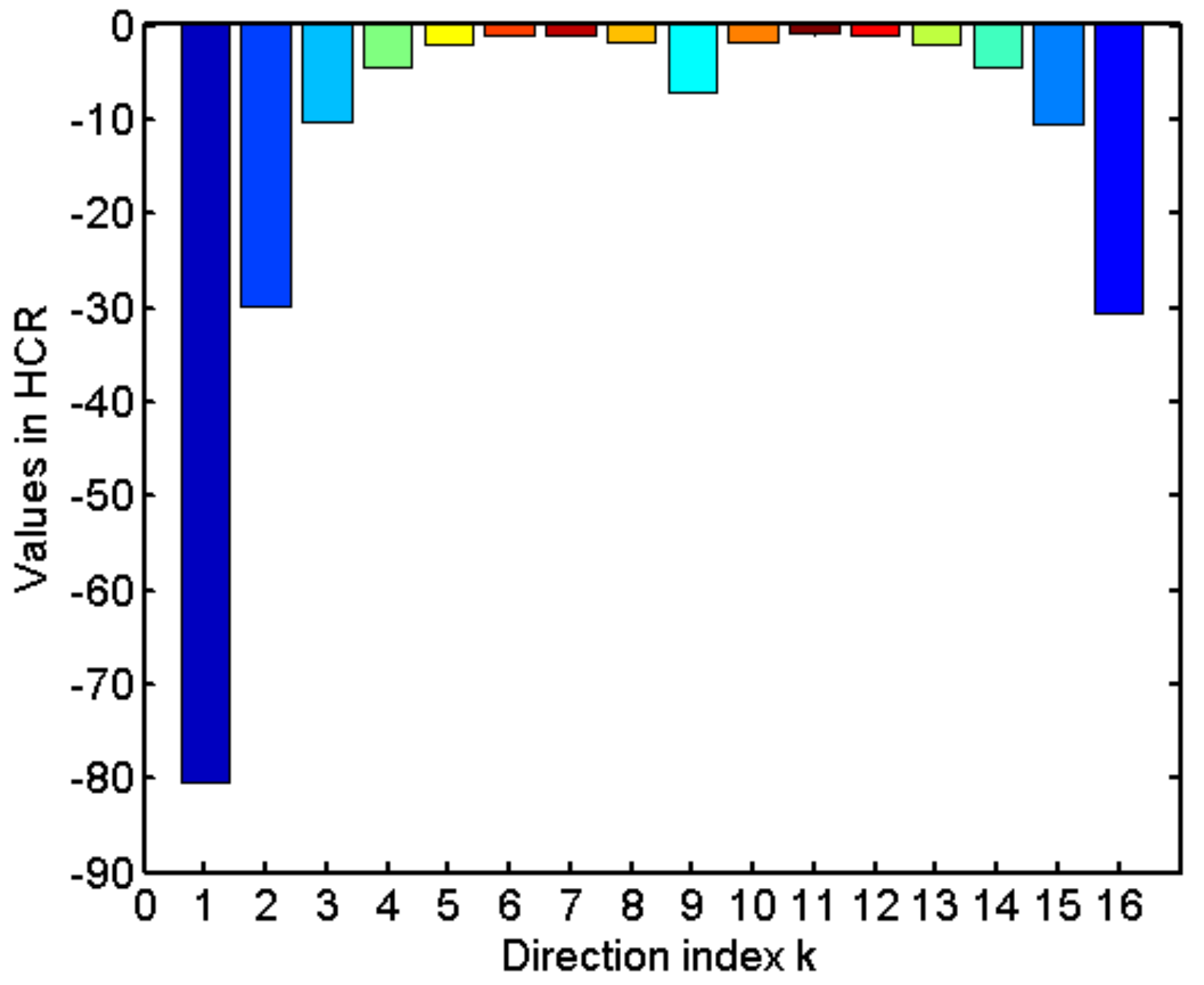}
	\caption{The values in each direction in ${{\bf{HCR}}}$.}
\end{figure}
\begin{figure}[tb]
	\centering
	\includegraphics[width=2in]{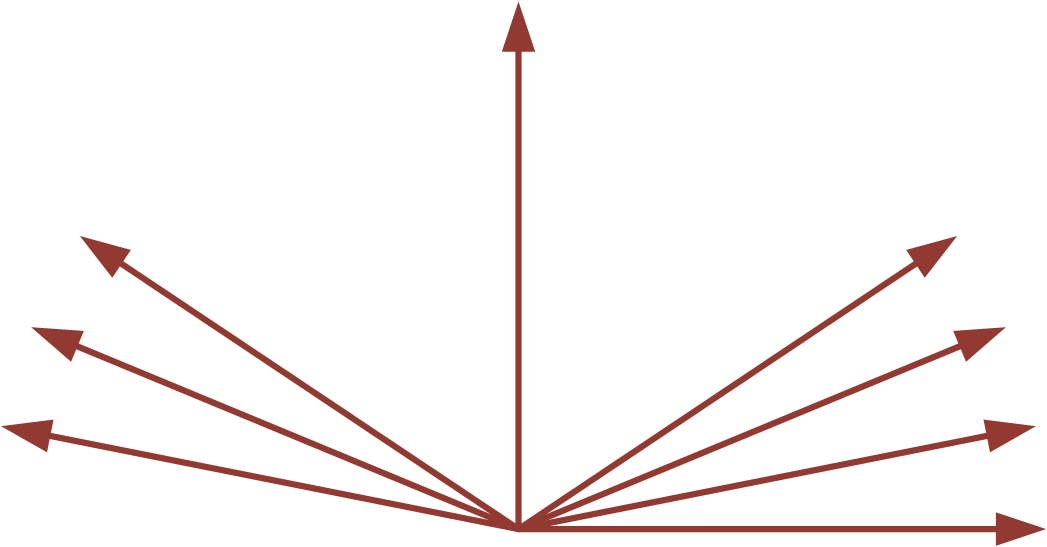}
	\caption{The selected directions according to smaller values in ${{\bf{HCR}}}$.}
\end{figure}

\begin{table}[tb]
	\caption{The values in ${{\bf{HCR}}}$ after sorting and the corresponding directions.}
	\label{table}
	\small
	\setlength{\tabcolsep}{3pt}
	\begin{tabular}{|c|cccccccc|}
		\hline
		\textbf{value} &	-80.41&	-30.63&	-29.76&	-10.62&	-10.20&	-7.171&	-4.57&	-4.48\\
		\hline
		\textbf{$\theta$}& 0&$\frac{{15\pi }}{{16}} $&$\frac{{\pi }}{{16}} $&$\frac{{7\pi }}{{8}} $&$\frac{{\pi }}{{8}} $&$\frac{{\pi }}{{2}} $&$\frac{{13\pi }}{{16}} $&$\frac{{3\pi }}{{16}} $\\
		\hline
		\textbf{value} &-2.14&	-2.12&	-1.87&	-1.72&	-1.14&	-1.03&	-1.00&	-0.90\\
		\hline
		\textbf{$\theta$} &$\frac{{3\pi }}{{4}} $&$\frac{{\pi }}{{4}} $&$\frac{{7\pi }}{{16}} $&$\frac{{9\pi }}{{16}} $&$\frac{{5\pi }}{{16}} $&$\frac{{11\pi }}{{16}} $&$\frac{{3\pi }}{{8}} $&$\frac{{5\pi }}{{8}} $\\	
		\hline
	\end{tabular}
	\label{tab1}
\end{table}

\subsection{Evaluation on the Use of Physiological Characteristic of finger vein}

To evaluate the effectiveness of the intention of physiological characteristic, we compared our HOPGR with several local feature-based methods that have a similar framework to both SEMI-FV and MMVBNU\_6000 databases. Table II tabulates the EERs using Gabor filter as a basic method for comparison. It is clear that HOPGR, HCGR, and HOL obtained better results than HOG, as they benefitted from the application of the Gabor filter. In HCGR and HOL, increasing the quantity of directions of the Gabor function improved the representation performance of the feature. When \textit{O}  was increased from 8 to 16, the EERs decreased in both HOL and HCGR. The time cost of HOPGR was compared with HCGR and HOLin MMVBNU\_6000 database. Our codes ran in MATLAB on a PC with a 3.60GHz CPU and an 8.00GB memory. HOL and HCGR were tested in  8 and 16 directions. Results are reported in Table II. It can be seen that  computational costs correspondingly increased in both HCGR and HOL when \textit{O} was increased from 8 to 16 . The proposed HOPGR with 8 directions obtained better performance and had less time cost than the HOL and HCGR with 16 directions. Results illustrate that for finger vein images with relatively fixed texture distribution patterns, setting effective directions can result in a more effective representation of the image.
\subsection{Comparison with the State-of-the-Art Methods}

To verify the validity of the proposed method, some state-of-the-art methods for finger vein recognition were selected for comparison in the MMVBNU\_6000 database. Comparison methods included GFSSR, k-means hashing-based method (KMHM) \cite{su2017finger}, DBC, iterative quantization-based method (ITQM) \cite{wang2017integration}, enhanced maximum curvature method with histogram of oriented gradient (EMC+HOG), Gabor+Tri-branch structure, and PG-Gabor. Performances are reported in Table III. Results of comparative methods as reported in the related references were used directly. Results show that HOPGR, our proposed method, can obtain better results than most state-of-the-art finger vein methods, including GFSSR, KMHM, DBC, ITQM, EMC+HOG, and a combination of primary and soft biometric traits. It can also achieve better performance than Gabor+Tri-branch structure, which is a Gabor filter-based method. HOPGR also exhibits a very competitive performance as compared with other Gabor filter-based methods, including Gabor+Tri-branch structure and PG-Gabor.  

Moreover, in our experiments, the HKPU database that provides prior physiological information is quite different from the MMVBNU\_6000 and SEMI-FV databases that are used for performance evaluation. The samples from these three databases are collected from completely different subjects, environments, and devices. So our method is theoretically universal to all finger vein images by using the statistical physiological characteristic of finger veins.

\begin{table}
	\caption{Comparison with the EERs({\%}) and computational time (ms) of the local feature-based methods that have the similar framework.}
	\label{table}
	\small
	\setlength{\tabcolsep}{3pt}
	\begin{tabular}{|p{125pt}|p{40pt}|p{30pt}|p{25pt}|}
		\hline
		\multirow{2}{*}{\textbf{Method}}&\textbf{SEMI-FV}&\multicolumn{2}{|c|}{\textbf{MMVBNU\_6000} }\\
		\cline{2-4}
		& \textbf{\textit{EER}}& \textbf{\textit{EER}}&\textbf{\textit{Time}} \\
		\hline
		Gabor filter \cite{kumar2011human}      & 1.67& 2.42 & *\\
		HOG \cite{dalal2005histograms}        &   	0.86&   1.54 & *\\
		HCGR \cite{lu2014finger} with 8 directions  &  0.80& 1.39& 85\\
		HCGR \cite{lu2014finger} with 16 directions  &  0.71& 1.07& 114\\   
		HOL \cite{Zhangliping2018} with 8 directions  &	0.74&   1.47 & 90\\
		HOL \cite{Zhangliping2018} with 16 directions  &	0.69& 1.00 & 120\\
		\textbf{Ours HOPGR with 8 directions}	&  \textbf{0.678} &  \textbf{0.70} &\textbf{90}\\
		\hline
	\end{tabular}
	\label{tab1}
\end{table}

\begin{table}
	\caption{Comparison with the EERs({\%}) of the State-of-the-Art Methods on MMVBNU\_6000.}
	\label{table}
	\small
	\setlength{\tabcolsep}{3pt}
	\begin{tabular}{|p{160pt}|p{30pt}|p{40pt}|}
		\hline
		\textbf{Method}& 
		\textbf{Year}& 
		\textbf{EER}\\
		\hline
		GFSSR \cite{xie2015intensity} & 2015 &1.5 \\
		KMHM \cite{su2017finger} & 2017 & 2.08\\
		DBC \cite{xi2017learning} & 2017& 2.48\\
		ITQM \cite{wang2017integration} & 2017 & 1.33\\
		EMC+HOG \cite{syarif2017enhanced} & 2017 & 1.79\\
		Gabor+Tri-branch structure \cite{yang2017tri} &2017& 1.14\\
		Combining primary and soft biometric traits \cite{kang2018noise}&2018 & 0.82\\
		PG-Gabor \cite{yang2019point} &2019& 0.71\\
		\textbf{Ours} &\textbf{2020}&  \textbf{0.70} \\
		\hline
	\end{tabular}
	\label{tab1}
\end{table}

\section{Conclusion}
This paper proposes a novel local feature descriptor named HOPGR that uses physiological characteristic of a finger vein. Theoretical analysis and extensive experimental results using the MMVBNU\_6000 and SEMI-FV databases clearly demonstrate the feasibility and effectiveness of the proposed method. However, although HOPGR demonstrates competitive performance, the physiological characteristic of the finger veins are used only to a preliminary degree in it. In the future, a more developed HOPGR will be set up in order to consider not only the directional distribution characteristic, but also the texture distribution characteristic of finger veins. It is reasonable to apply the proposed method of direction setting that relies on physiological characteristic to other Gabor-based methods.

\section*{Acknowledgment}

This work was supported by the National Nature Science Foundation of China, grant no. 61901436.

\bibliographystyle{IEEEtran}
\bibliography{IEEEabrv,reference.bib}

\end{document}